%% file: main.tex
\PassOptionsToPackage{table}{xcolor}
\documentclass[sigconf]{acmart}

\AtBeginDocument{%
  }

\usepackage{hyperref}       
\usepackage{url}            
\usepackage{booktabs}       
\usepackage{amsfonts}       
\usepackage{nicefrac}       
\usepackage{microtype}      
\usepackage[table]{xcolor} 
\usepackage{makecell}
\usepackage{enumitem}
\usepackage{array}
\usepackage{wrapfig}
\usepackage{rotating}
\usepackage{amsmath}
\usepackage{subfigure}
\usepackage{xcolor}
\usepackage{tcolorbox}
\tcbuselibrary{most}
\usepackage{colortbl}
\usepackage{caption}
\usepackage{subcaption}
\usepackage{tabularx}
\usepackage{multirow}
\usepackage{twemojis}
\usepackage{titlesec}
\usepackage{listings}
\usepackage{xcolor}
\usepackage{xspace}
\titleformat{\section}{\normalfont\Large\bfseries\MakeUppercase} 
  {\thesection}{1em}{}

\definecolor{myshadow}{rgb}{0.901,0.901,0.901}
\definecolor{darkgreen}{rgb}{0,0.5,0} 
\definecolor{purple}{rgb}{1,0,1} 
\definecolor{todocolor}{rgb}{0.9,0.1,0.1} 
\definecolor{fixcolor}{rgb}{0.1,0.7,0.3} 
\definecolor{wycolor}{rgb}{0.9,0.1,0.1} 
\definecolor{hycolor}{rgb}{0.7,0.7,0.3} 
\definecolor{ghcolor}{rgb}{0.79, 0.63, 0.86} 
\definecolor{ashgrey}{rgb}{0.7, 0.75, 0.71}
\definecolor{grey}{rgb}{0.6,0.6,0.6}
\definecolor{lightblue}{rgb}{0.76, 0.82, 0.94}
\definecolor{czqcolor}{HTML}{00aeec}

\definecolor{lightred}{RGB}{254,246,245}
\definecolor{lightgreen}{RGB}{240,249,237}
\definecolor{lightblue}{RGB}{238,244,250}
\definecolor{lightcyan}{RGB}{10,110,150}

\newcommand{\code}[1]{\textasciigrave#1\textasciigrave}
\newcommand{\mypara}[1]{\smallskip \noindent\textbf{#1} \xspace}
\newcommand{\codemore}[1]{\textasciigrave\textasciigrave\textasciigrave#1\textasciigrave\textasciigrave\textasciigrave}

\newcommand{\method}{\textsc{CGBridge}\xspace}%

\newtcolorbox{promptbox}{
    enhanced,
    colback=gray!10!white,
    colframe=gray!80!black,
    fonttitle=\bfseries,
}

\copyrightyear{2025}
\acmYear{2025}
\setcopyright{acmlicensed}
\acmConference[KDD '25] {Proceedings of the 31st ACM SIGKDD Conference on Knowledge Discovery and Data Mining V.1}{August 3--7, 2025}{Toronto, ON, Canada.}
\acmBooktitle{Proceedings of the 31st ACM SIGKDD Conference on Knowledge Discovery and Data Mining V.1 (KDD '25), August 3--7, 2025, Toronto, ON, Canada}
\acmISBN{979-8-4007-1245-6/25/08}
\acmDOI{	https://doi.org/10.1145/3690624.3709256}
\renewcommand\footnotetextcopyrightpermission[1]{} 

\begin{document}
\title{Bridging Code Graphs and Large Language Models for Better Code Understanding}
\author{Zeqi Chen$^{1}$, Zhaoyang Chu$^2$, Yi Gui$^2$, Feng Guo$^1$\\Yao Wan$^{2\dagger}$, and Chuan Shi$^{1\dagger}$}
\thanks{\\$\dagger$ Corresponding author.}
\affiliation{$^1$Beijing University of Posts and Telecommunications \country{China} \\$^2$Huazhong University of Science and Technology \country{China}
\\\twemoji{e-mail} {chenzeqi@bupt.edu.cn, shichuan@bupt.edu.cn}  
}

\renewcommand{\shortauthors}{Zeqi Chen et al.}


\begin{abstract}

Large Language Models (LLMs) have demonstrated remarkable performance in code intelligence tasks such as code generation, summarization, and translation. 
However, their reliance on linearized token sequences limits their ability to understand the structural semantics of programs.
While prior studies have explored graph-augmented prompting and structure-aware pretraining, they either suffer from prompt length constraints or require task-specific architectural changes that are incompatible with large-scale instruction-following LLMs.
To address these limitations, this paper proposes \textbf{\method}, a novel plug-and-play method that enhances LLMs with \underline{C}ode \underline{G}raph information through an external, trainable \underline{\textsc{Bridge}} module. 
\method first pre-trains a code graph encoder via self-supervised learning on a large-scale dataset of 270K code graphs to learn structural code semantics. 
It then trains an external module to bridge the modality gap among code, graph, and text by aligning their semantics through cross-modal attention mechanisms.
Finally, the bridge module generates structure-informed prompts, which are injected into a frozen LLM, and is fine-tuned for downstream code intelligence tasks.
Experiments show that \method achieves notable improvements over both the original model and the graph-augmented prompting method. Specifically, it yields a 16.19\% and 9.12\% relative gain in \textit{LLM-as-a-Judge} on code summarization, and a 9.84\% and 38.87\% relative gain in \textit{Execution Accuracy} on code translation. 
Moreover, \method achieves over 4× faster inference than LoRA-tuned models, demonstrating both effectiveness and efficiency in structure-aware code understanding \footnote{Code are available at: https://github.com/OmniJax/CGBridge.}.

\end{abstract}





\maketitle


\section{Introduction}

Large Language Models (LLMs) have revolutionized software engineering by excelling in tasks such as code generation, summarization, and translation~\cite{wang2021codet5,copilot-codex,transformerbased-codesum,bart, transcoder,autounittest4codetrans}, as demonstrated by advanced code-focused models like Code Llama~\cite{codellama}, Qwen2.5-Coder~\cite{qwen2.5coder}, DeepSeek-Coder~\cite{guo2024deepseekcoder}, and commercial tools such as Copilot~\cite{copilot-codex} and Cursor~\cite{cursor2024}.
This success can be attributed to large-scale pre-training on source code, which allows models to learn and generalize the intrinsic patterns and semantics of programs.

Despite the impressive success of the current pre-training paradigm, LLMs often process code as plain text by linearizing it into flat token sequences, which overlooks its inherent structure~\cite{COMEX-structualblindness, HierarchyNet}.
Yet, source code naturally encodes rich syntactic and semantic information through structures like abstract syntax trees (ASTs)~\cite{AST-code2vec, AST-code2seq}, control flow graphs (CFGs)~\cite{CFG-androidapp, CFG-CNNCFG}, and data flow graphs (DFGs)~\cite{guo2021-graphcodebert}.
As a result, training LLMs only on surface-level code text limits their ability to capture program semantics and handle tasks that require comprehending program logic.

\begin{figure}[!t]
    \centering
    \includegraphics[width=0.99\linewidth]{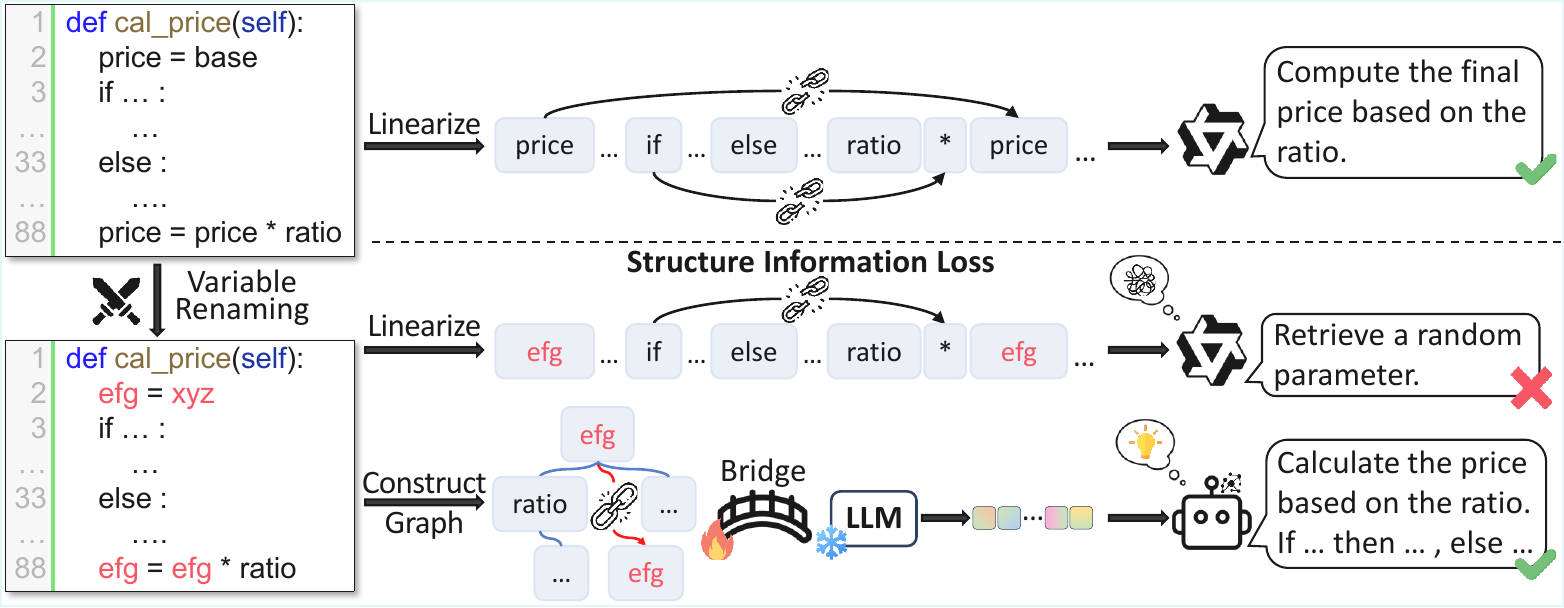}
    \vspace{-1em}
    \caption{A motivating example.
    }
    \label{fig:challenges}
    \vspace{-2em}
\end{figure}

\autoref{fig:challenges} illustrates a motivating example highlighting the inherent weaknesses of current code LLMs. 
Qwen2.5-Coder successfully summarizes the original snippet as ``\textit{compute the final price based on the ratio}''. Yet, after the variable \texttt{price} is renamed to \texttt{efg}, the generated summary becomes irrelevant, \emph{e.g.}, ``\textit{retrieve a random parameter}''. 
This suggests that models relying solely on linearized token sequences fail to capture essential structural and semantic information, making them vulnerable to minor textual perturbations like variable renaming. 
These observations motivate us to investigate the key question: \textit{Can LLMs be effectively bridged with structural code semantics to achieve deeper program understanding?}

Currently, many efforts have attempted to help LLMs understand graph structures by serializing them into text sequences for prompt-based inference~\cite{zhao2023graphtext, guo2023gpt4graph, liu2023eval_llms_on_graphs, liu2024code_prompting_vul}. While showing promise in general domains, these methods often inflate prompt length and dilute structural knowledge—limitations that particularly hinder performance on code understanding tasks.
Moreover, several structure-aware learning methods—such as GraphCodeBERT~\cite{guo2021-graphcodebert} and UniXCoder~\cite{guo2021-unixcoder}—have been proposed to incorporate code graph information during model pre-training. 
However, these methods are primarily designed for supervised fine-tuning of small-scale models on specific downstream tasks and typically rely on task-specific architectural modifications (\textit{e.g.}, adding decoders) for generation, lacking compatibility with end-to-end instruction following.
Extending such pre-training and fine-tuning paradigms to large-scale LLMs is often impractical due to prohibitive computational costs.

To overcome the above limitations, we introduce \textbf{\method}, a plug-and-play method that augments LLMs with \underline{C}ode \underline{G}raphs for code intelligence tasks via an external, trainable \underline{\textsc{Bridge}} module that learns structural semantics from code.
To implement \method, we construct a large-scale dataset of approximately 270K code property graphs comprising heterogeneous structural components such as ASTs, CFGs, and DFGs.
We then pre-train a code graph encoder on this dataset using self-supervised learning to capture rich structural information.
Subsequently, we train a bridge module to align representations across code, graph, and text modalities through cross-modal attention, enabling semantic integration between structured and unstructured inputs.
Finally, we utilize the bridge module to generate structure-informed prompts, which are injected into a frozen LLM, and fine-tune the bridge for adaptation to downstream tasks such as code summarization and code translation.
Our proposed \method is fully compatible with existing LLMs and can be seamlessly integrated without modifying the underlying pre-trained models.

In summary, our key contributions can be summarized as follows:
\begin{itemize}[leftmargin=*,nosep]

    \item 
    \textbf{New Dataset.}
    We curate a comprehensive code graph dataset comprising approximately 270K samples, each annotated with heterogeneous code entities, including abstract syntax trees and control/data flows. This dataset serves as a valuable resource for training and evaluating models on structural code understanding.

    \item 
    \textbf{Novel Approach.}
    We introduce a novel plug-and-play framework, \method, which enhances large language models by seamlessly integrating an external, trainable bridge module designed to learn and inject code structural semantics, without altering the underlying pre-trained model.
    

    \item
    \textbf{Comprehensive Evaluation.}
    We conduct extensive experiments on three widely used families of code-specific LLMs, namely Qwen2.5-Coder~\cite{qwen2.5coder}, Code Llama~\cite{codellama}, and DeepSeek-Coder~\cite{guo2024deepseekcoder}.
    The results on two representative code understanding tasks (\emph{i.e.}, code summarization and translation) demonstrate the effectiveness of \method in enhancing LLMs' understanding of code structural semantics.

\end{itemize}


\begin{figure}[!t]
    \centering
    \includegraphics[width=0.99\linewidth]{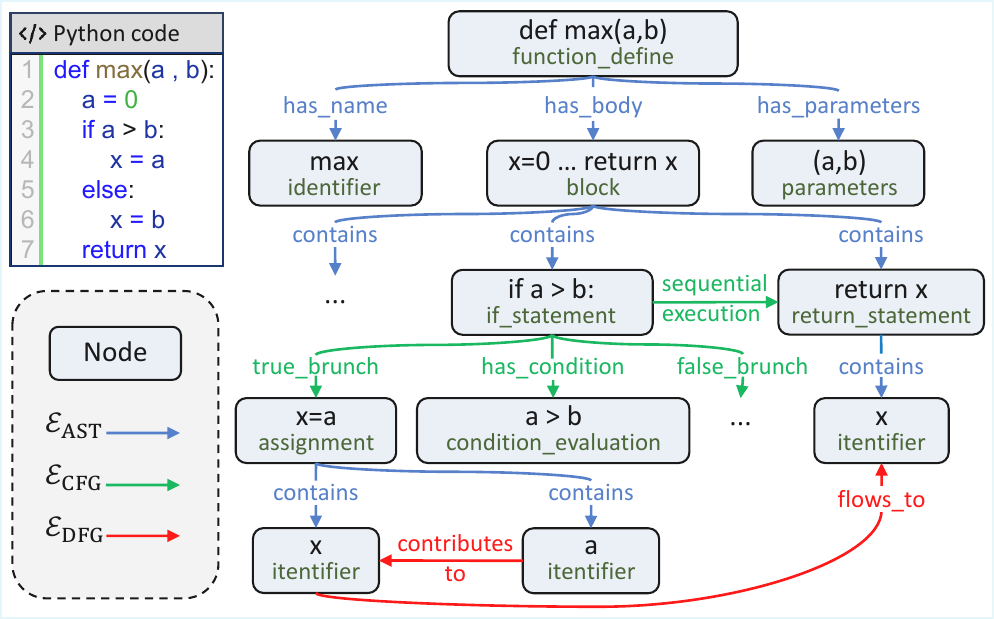}
    \vspace{-1em}
    \caption{A Python function (left) and its corresponding Code Property Graph (CPG, right).}
    \label{fig:CodeGraphRepresentation}
    \vspace{-4mm}
\end{figure}


\section{Preliminaries}

\subsection{Code Property Graph}

Code Property Graph (CPG) is a unified, graph-based representation of source code widely used in program analysis~\cite{CPG-Yamaguchi, repllvmirincpg, TAILOR-CPGNN, wi2022hiddencpg}.
It provides a holistic view of complex code entities by integrating various structural components, including abstract syntax trees (ASTs), control flow graphs (CFGs), and data flow graphs (DFGs).
AST edges encode the grammatical hierarchy and composition of code elements, such as variables and statements~\cite{AST-code2seq, AST-code2vec}. 
CFG edges represent the control flows among program statements~\cite{CFG-CNNCFG, CFG-androidapp}.
DFG edges track the dependencies between variable definitions and their uses~\cite{guo2021-graphcodebert}.


\autoref{fig:CodeGraphRepresentation} illustrates a simplified CPG for a Python function.
AST edges (marked in blue) capture the syntactic structure by linking the \texttt{function\_define} node to its \texttt{name}, \texttt{parameters}, and \texttt{body}, and by recursively connecting statements and expressions within the function.
CFG edges (marked in green) encode the control flow, explicitly representing the \texttt{true\_branch} and \texttt{false\_branch} of the conditional statement.
DFG edges (marked in red) track data dependencies, indicating how values (\emph{e.g.}, \texttt{a}) are propagated and contribute to the assignment and subsequent use of variables (\emph{e.g.}, \texttt{x}) throughout the function.
The construction details of CPG are provided in~\autoref{sec_code_graph_dataset}.

\subsection{Notations and Problem Formulation}

Before presenting the proposed approach, we define the key notations used in this paper.
Let $C$ represent a code snippet, which is modeled as a heterogeneous graph, the CPG, formally defined as $G=(\mathcal{V}, \mathcal{E}, \mathcal{T}^v, \mathcal{T}^e)$.
Here, $\mathcal{V}$ denotes the set of nodes representing syntactic code elements, with a node type mapping function $\phi\,:\, \mathcal{V} \to \mathcal{T}^v$.
Each node $v_i \in \mathcal{V}$ is associated with a node type $t^v_i = \phi(v_i) \in \mathcal{T}^v$, which defines the specific type of the corresponding code element (\emph{e.g.}, \texttt{identifier} or \texttt{assignment}).
$\mathcal{E}$ denotes the set of edges, with an edge type mapping function $\psi\,:\,\mathcal{E} \to \mathcal{T}^e$.
Each edge $e_{ij} = (v_i, v_j) \in \mathcal{E}$ is associated with a relation $t^e_{ij} = \psi(e_{ij}) \in \mathcal{T}_e$, which indicates the relationship between nodes $v_i$ and $v_j$ (\emph{e.g.}, the AST edge \texttt{contains}, the CFG edge \texttt{has\_condition}, or the DFG edge \texttt{flows\_to}).
A detailed enumeration of node and edge types is provided in~\autoref{sec:nodes_edges_textual_attributes}.


\mypara{Problem Statement.}
Given a code snippet $C$ and its corresponding CPG representation $G$, our goal is to address the code intelligence task (\emph{e.g.}, code summarization or translation) by generating an appropriate answer text $T_A = (t_1, t_2, \ldots, t_m)$ conditioned on a natural language instruction $I$.
Here, $t_i$ denotes the $i$-th token in the generated sequence of length $m$.
Formally, the objective is to optimize the model parameters $\theta$ by maximizing the conditional probability of generating $T_A$ given the input triplet $(C, G, I)$:
\begin{equation}
    \max _{\theta} P\left(T_A \mid C, G, I; \theta\right)=\max _{\theta} \prod_{i=1}^{m} P\left(t_i \mid t_{<i}, C, G, I; \theta\right)\,,
\end{equation}
where $t_{<i}$ denotes the sequence of tokens prior to token $t_i$.


\section{\method: the Proposed Method}

\begin{figure*}[tp]
    \centering
     \includegraphics[width=1.0\linewidth]{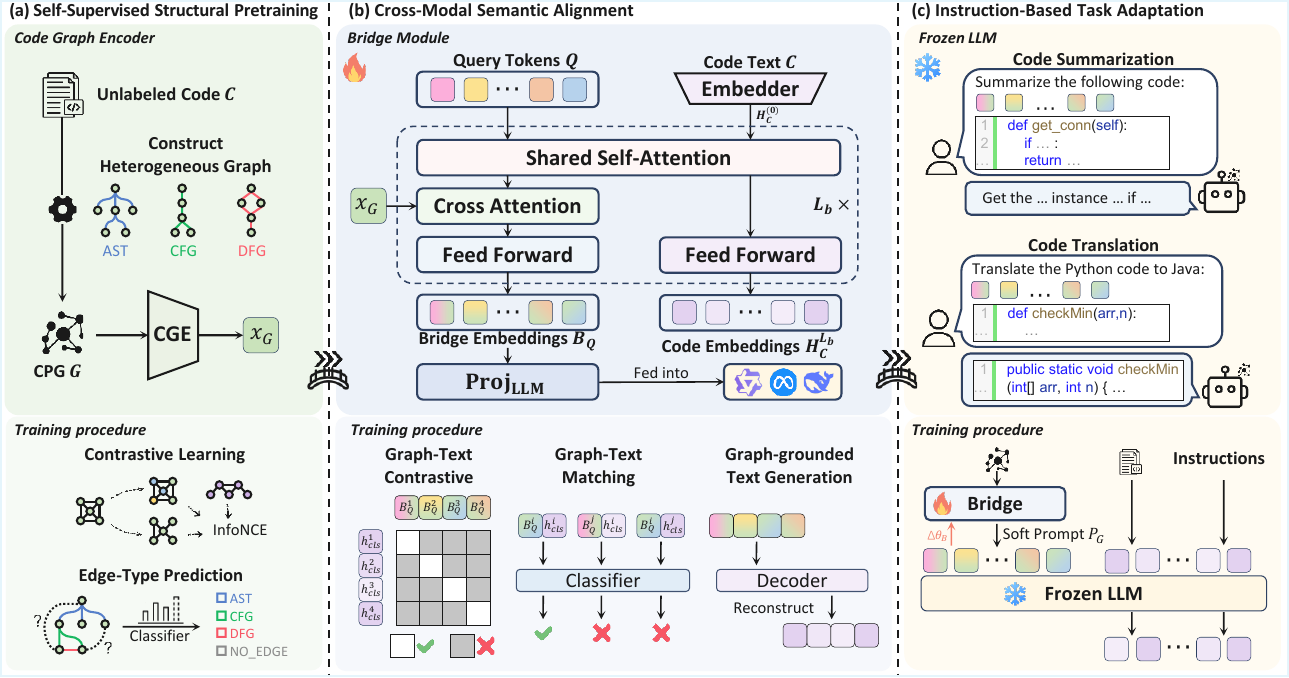}
     \vspace{-2.5em}
    \caption{Overview of the \method framework.}
    
    \label{fig: Bridge Overview}
    \vspace{-4mm}
\end{figure*}

We propose \textbf{\method}, a novel plug-and-play approach for incorporating \underline{C}ode \underline{G}raph information into LLMs via a trainable \underline{\textsc{Bridge}} module that learns code structural semantics.
As illustrated in~\autoref{fig: Bridge Overview}, \method employs a sequential three-stage training procedure:
\textbf{(a) Self-Supervised Structural Pretraining.}
\method first pre-trains a general encoder on large-scale code graphs to capture rich structural information using two self-supervised objectives, \emph{i.e.}, graph-level contrastive learning and edge type prediction.
\textbf{(b) Cross-Modal Semantic Alignment.}
Next, \method trains an external module to bridge the modality gap among code, graph, and text by aligning their representations via three complementary objectives \emph{i.e.}, graph-text contrastive learning, graph-text matching, and graph-grounded text generation.
\textbf{(c) Instruction-Based Task Adaptation.}
Finally, \method leverages the bridge module to inject learned code structural semantics into the frozen LLM and fine-tunes it for adaptation to downstream code intelligence tasks, \emph{i.e.}, code summarization and code translation.
We will elaborate on each component of \method in the following.

\subsection{Self-Supervised Structural Pretraining}
\subsubsection{\textbf{Code Graph Encoder.}}
\label{sec:CodeGraphEncoder}


We model source code as a heterogeneous CPG $G$, and leverage a code graph encoder (CGE) to transform the code graph into a graph-level embedding $x_G$.
We implement the encoder using an $L_g$-layer Graph Neural Network (GNN) that operates via message passing.
Each GNN layer updates a node's representation by aggregating edge-conditioned messages from its neighborhood. 
This process can be formally expressed as:
\begin{equation}
    h_i^{(l)} = \mathrm{UPDATE} \left( h_i^{(l-1)}, \mathop{\mathrm{AGGREGATE}}_{j \in \mathcal{N}(i)} \left( \left\{ \left( h_j^{(l-1)}, e_{ji} \right) \right\} \right) \right)\,.
\end{equation}
Here, $\mathcal{N}(i)$ is the set of neighbors of node $i$. The $\mathrm{AGGREGATE}(\cdot)$ and $\mathrm{UPDATE}(\cdot)$ functions are the core of the layer. To handle our graph's heterogeneity, these functions are critically conditioned on the edge features $e_{ji}$, in addition to the node features, which allows the CGE to distinguish between different relationship types (\emph{e.g.}, syntactic vs. data-flow). After $L_g$ layers of message passing, we use a $\mathrm{READOUT}(\cdot)$ to output the final node embeddings into $x_G$.


\subsubsection{\textbf{Training Procedure.}}
To ensure the CGE can produce meaningful graph representations, we first pre-train it on a large corpus of unlabeled code graphs. The training is guided by the following two self-supervised objectives.

\mypara{Graph-Level Contrastive Learning.} 
To learn a robust global representation, we adopt a contrastive learning strategy~\cite{Hassani-multiviewRepresentationLearningonGraphs,You-GraphConstrastiveLearningwithAug}. For each graph, we generate two augmented views (\emph{e.g.}, via node feature masking and edge dropping) and train the CGE to maximize their embedding similarity while minimizing it with other graphs in the batch, using an InfoNCE loss~\cite{vanden-infoNCE}.

\mypara{Edge-Type Prediction.} 
To learn fine-grained relational semantics, a multiclass classifier is trained on the embeddings of node pairs to predict the edge type between them. We include negative sampling by creating non-existent edges, which the model must classify as a distinct ``NO\_EDGE'' type \cite{grover-node2vec-negasample,hamilton-graphsage-negasample}.

The loss for Stage 1 is a weighted sum of the two objectives:
\begin{equation}
\mathcal{L}_\text{Stage1} = \lambda_{cl} \cdot \mathcal{L}_\text{cl} + \lambda_{edge} \cdot \mathcal{L}_\text{edge} \,.
\end{equation}

\subsection{Cross-Modal Semantic Alignment}
\subsubsection{\textbf{Bridge Module.}}
\label{sec:bridge}
The bridge module addresses the modality gap between the graph embeddings and the LLM's sequential input space. 
Architecturally, it is a $L_b$-layer Transformer~\cite{Vaswani-attentionisallyouneed}.
Inspired by recent works in multi-modal learning \cite{li-BLIP2}, the module employs a set of $N_q$ learnable query tokens, $Q \in \mathbb{R}^{N_q \times d_{\text{model}}}$, to jointly model information from both the graph and text. 
To bridge the gap, the module accepts two key inputs: the graph's structural embedding, $x_G$, and the code's textual embeddings $H_C$. The latter is produced from the raw code $C$ by the module's text embedder.
The $Q$ then interact with these inputs through distinct attention mechanisms within each of the $L_b$ layers.
The three steps are as follows:

\mypara{1) Multi-Modal Fusion via Shared Self-Attention.}
$Q^{(l-1)}$ and $H_C^{(l-1)}$ from the previous layer are concatenated and processed by a \textbf{shared self-attention layer}, allowing them to exchange information. For the first layer ($l=1$), the inputs are the initial learnable queries $Q$ and code text embeddings $H_C$:
\begin{equation}
        [Q_{sa}^{(l)}, H_{C,sa}^{(l)}] = \text{Self-Attn}([Q^{(l-1)}, H_C^{(l-1)}])\,.
\end{equation}

\mypara{2) Graph Information Extraction via Cross-Attention.} 
$Q_{sa}^{(l)}$ from the self-attention step then acts as the query to extract relevant structural information from $x_G$ via \textbf{cross-attention}:
\begin{equation}
Q_{cross}^{(l)} = \text{Cross-Attn}(Q = Q_{sa}^{(l)}, K = x_G, V = x_G)\,.
\end{equation}

\mypara{3) Feed-Forward Network.} 
Finally, the outputs from the attention layers are processed through separate Feed-Forward Networks (FFN) to further refine their representations.
\begin{equation}
Q^{(l)} = \text{FFN}_Q(Q_{cross}^{(l)}), \quad H_C^{(l)} = \text{FFN}_C(H_{C,sa}^{(l)})\,.
\end{equation}
The final output of the Bridge Module is the set of contextualized query embeddings from the last layer, $B_Q = Q^{(L_b)} \in \mathbb{R}^{N_q \times d_\text{model}}$. 
To interface with the downstream LLM, these embeddings are linearly projected into a soft prompt  $P_G = \text{Proj}_{\text{LLM}}(B_Q)$, which will later be used to guide task-specific generation.


\subsubsection{\textbf{Training Procedure.}}
This stage aligns the representations of code graph-text pairs $(x_G,C)$ within the Bridge module, independent of the final LLM (\autoref{fig: Bridge Overview}(b)). 
Following established multi-modal learning methodologies \cite{li-BLIP2,liu2023llava}, we employ three complementary objectives:

\mypara{Graph-Text Contrastive Learning (GTC).} 
This objective aligns global representations of graphs and texts.
Given a batch of M pairs, let $\mathbf{B_Q} = \{B_Q^i\}_{i=1}^M$ denote the graph-query embeddings from the Bridge module.
Critically, let $\mathbf{h_{cls}} = \{h_{cls}^i\}_{i=1}^M$ denote the text-only representations, where each $h_{cls}^i$ is the \texttt{[CLS]} token of $H_C^{(L_b)}$ for its sample.
The symmetric InfoNCE loss \cite{vanden-infoNCE} encourages aligned pairs $(B_Q^i,h_{cls}^i )$ to be similar and contrasts them against all other negative pairs:
\begin{equation}
    \label{eq:gtc_loss}
    \mathcal{L}_{\text{GTC}} = \frac{1}{2} \left(\text{InfoNCE} ( \mathbf{B_Q}, \mathbf{h_{{cls}}},\tau ) + \text{InfoNCE}( \mathbf{h_{{cls}}} , \mathbf{B_Q},\tau) \right) \,.
\end{equation}
The similarity within InfoNCE function is computed via maximum cosine similarity over all $N_q$ query tokens, and the resulting scores are then scaled by a learnable temperature parameter $\tau$ (detailed in \autoref{sec:infonce}).

\mypara{Graph-Text Matching (GTM).} 
This objective trains a binary classifier $\phi_{\text{gtm}}$, to predict whether a graph-text pair is a true match. 
Given a pair $(B_Q^i,h_{cls}^i)$, the predicted probability $p^i=\phi_{\text{gtm}}(B_Q^i,h_{cls}^i)$.
The loss is standard cross-entropy averaged over $M$ samples, with hard negative mining \cite{HardNegatives} applied to construct challenging non-matching pairs: 
\begin{equation}
    \mathcal{L}_{\text{GTM}} = -\frac{1}{M} \sum_{i=1}^{M} \left( y^i \log(p^i) + (1-y^i) \log(1-p^i) \right) \,.
\end{equation}

\mypara{Graph-grounded Text Generation (GTG).}
This objective encourages $B_Q$ to encapsulate a comprehensive summary of the graph by training the model to reconstruct the $C$.
During this autoregressive decoding, each token attends to $B_Q$, grounding the generation in the structural representation.
The loss is the average negative log-likelihood over $M$ samples, where the model predicts each token $t_{C,j}^i$ of the $i$-th sequence $t_C^i$ conditioned on its preceding tokens $ t_{C,<j}^i$ and the corresponding $B_Q^i$:
\begin{equation}
    \mathcal{L}_{\text{GTG}} = -\frac{1}{M} \sum_{i=1}^{M} \sum_{j=1}^{|t_C^i|} \log P\left(t_{C,j}^i \mid t_{C,<j}^i, B_Q^i\right) \,.
\end{equation}
The loss for this stage is the sum of the three objectives: 
\begin{equation}
\mathcal{L}_\text{Stage2}=\mathcal{L}_\text{GTC}+\mathcal{L}_\text{GTM}+\mathcal{L}_\text{GTG} \,.
\end{equation}

\subsection{Instruction-Based Task Adaptation}
\subsubsection{\textbf{Frozen LLM.}}
\label{sec:FrozenLLM}
\method utilizes a pre-trained, domain-specialized code LLM as its final inference engine (\autoref{fig: Bridge Overview}(c)). 
A core tenet of our methodology is that the LLM's parameters remain entirely frozen. 
This deliberate design choice allows us to leverage the powerful, existing knowledge of the foundation model for downstream tasks in an end-to-end manner, while all task-specific adaptation is efficiently handled by our efficient Bridge module. 
This approach avoids the prohibitive computational costs associated with fine-tuning the LLM itself.


\subsubsection{\textbf{Training Procedure.}}
Following cross-modal alignment, this final stage adapts the Bridge module to downstream tasks under a frozen LLM setup (\autoref{fig: Bridge Overview}(c)).
Given a code snippet, the Bridge processes its graph embedding $x_G$ to produce $B_Q$, which is projected to a soft prompt $P_G=\text{Proj}_\text{LLM}(B_Q)$.
This $P_G$ is prepended to the task-specific instruction $I$  and and the code $C$, forming the composite input, $C_\text{LLM}=[P_G; I; C]$.
The frozen LLM is then prompted to autoregressively generate the ground-truth answer sequence $T_A$ based on $C_\text{LLM}$. The training is supervised, optimizing only the Bridge module's parameters $\theta_B$, by minimizing the negative log-likelihood of the ground-truth tokens:
\begin{equation}
\mathcal{L}_{\text{Stage3}}(\theta_B) = - \sum_{i=1}^{|T_A|} \log P_{\text{LLM}}(t_{i} \mid t_{<i}, C_{\text{LLM}}) \,.
\end{equation}

\section{Experiments}\label{sec: exp}

\input{tab/MainResults}

\subsection{Code Graph Dataset}
\label{sec_code_graph_dataset}

We conduct comprehensive evaluations on two large-scale, publicly available datasets: \textit{CodeSearchNet (Python subset)} \cite{codesearchnet} for code summarization and \textit{XLCoST (Python-to-Java)} \cite{zhu2022xlcost} for code translation. 
To establish a foundation for our experiments, we refined and constructed a large-scale code graph dataset comprising approximately \textbf{270K samples} for our two downstream tasks. This process involved two main steps: data refinement and graph construction. The detailed implementations are provided in \autoref{sec:detail_of_codegraphdataset}.


\mypara{Dataset Refinement.}    
    To improve the quality of the source data, we performed task-specific refinement. For the \textit{CodeSearchNet}, 
    we removed comments in the code and utilized GPT-4 to refine the ground-truth summaries.
    For the \textit{XLCoST}, we processed and formatted the code snippets, ensuring their syntactic correctness and executability.

\mypara{Graph Construction.}
    Following the refinement, we construct a CPG for each snippet. 
    We begin by parsing source code using tree-sitter\footnote{\url{https://tree-sitter.github.io/tree-sitter}}, 
    where each node corresponds to a named AST element and is labeled with a predefined type (\emph{e.g.}, \texttt{assignment}, \texttt{identifier}).
    For textual representation, high-level statement nodes retain only the first line of their source text (\emph{e.g.}, the condition in an \texttt{if} statement), while other nodes (\emph{e.g.}, \texttt{expressions}, \texttt{identifiers}) retain their full span. 
    Edges in the CPG encode structural and semantic relations and are categorized into AST, CFG, and DFG types.
    AST edges capture syntactic hierarchy (\emph{e.g.}, \texttt{has\_name}, \texttt{contains}).
    CFG and DFG edges are both derived from the static analysis of the AST: 
    CFG edges encode control-flow semantics (\emph{e.g.}, \texttt{sequential\_execution}, \texttt{true\_branch}) by analyzing statement sequences and control structures, while DFG edges model data dependencies (\emph{e.g.}, \texttt{flows\_to}, \texttt{contributes\_to}) based on variable definition and usage patterns.
    Each edge is also assigned a textual attribute indicating its semantic role.
    Finally, all node and edge textual attributes are encoded into dense feature vectors using a pre-trained code encoder (\emph{e.g}., UniXCoder), and the resulting graph is stored and processed using the PyG \cite{PyG} for downstream GNN computation.

\subsection{Experimental Setup}
\mypara{Downstream Tasks.} 
We evaluate \method on two representative code generation tasks that require deep structural understanding: \textit{Code Summarization}, which involves identifying control flows, data dependencies and functional roles to generate accurate natural language descriptions of code functionality;
\textit{Code Translation}, which re-expresses the same logic in a different language while preserving semantics, relies heavily on the accurate modeling of abstract syntax trees and complex program structures.
These tasks serve as strong testbeds for \method's to unify structural comprehension with generative quality. 

\mypara{Metrics.} 
Our evaluation protocol combines lexical and semantic similarity, reference-free assessment, and functional correctness. Specifically, for summarization, we adopt \textit{METEOR} \cite{meteor} and \textit{ROUGE-L} \cite{lin-2004-rouge} for lexical similarity, \textit{BERTScore} \cite{BERTScore} and \textit{Sentence-BERT cosine similarity} \cite{Sentence-BERT}  for semantic similarity and the \textit{LLM-as-a-Judge} paradigm \cite{{Wang2025CanllmReplaceHumanEva,wu2024llmserveasEvaluator4codesum,kang2025memoryos,tong2024codejudgeevaluatingcodegeneration,gu2025surveyllmasajudge}}, where an LLM rates generated summaries from 0-4 on Coherence, Consistency, Fluency, and Relevance, which are then aggregated into an overall quality rating. For translation, we use \textit{CodeBLEU} \cite{CodeBLEU}, Syntax Match, and Dataflow Match to evaluate structural alignment, with \textit{Execution Accuracy} serving as the primary metric for functional correctness of the translated code.

\mypara{Baselines.} 
We compare \method against the following three baselines to demonstrate its effectiveness: 
\begin{itemize}[leftmargin=*,nosep]
    \item \textbf{Zero-shot.} Base code LLMs without task-specific fine-tuning.
    \item \textbf{+LoRA.} Code LLMs fine-tuned with LoRA \cite{lora}, a powerful and widely-adopted Parameter-Efficient Fine-Tuning (PEFT) method, which provides a primary text-only adaptation baseline.
    \item \textbf{+GraphText.} 
    Following prior work \cite{zhao2023graphtext,nlgraph}, we convert code graphs into textual descriptions (listing nodes then edges) as in-context information for LLMs.
\end{itemize}
The detailed implementations are referred to Appendixes~\ref{sec:prompt},~\ref{sec:implementation_details} and~\ref{sec:hyperparameters}.

\subsection{Overall Performance}
\autoref{tab:MainResults} presents the main experimental results of our proposed \method framework on two downstream tasks. 

\mypara{Performance on Code Summarization.}
The results reveal a consistent trend across all models. The +GraphText baseline exhibits limited effectiveness and occasionally underperforms the Zero-shot baseline, indicating that long serialized graph descriptions may introduce noise and hinder the LLM’s ability to exploit structural information. 
In comparison with the strong +LoRA baseline, our +\method shows a consistent advantage. While +LoRA remains competitive on lexical overlap metrics such as \textit{ROUGE-L}, \method consistently achieves superior performance on semantic metrics, including \textit{BERTScore} and \textit{LLM-as-a-judge} evaluations, demonstrating its capability to generate summaries with higher semantic fidelity and overall quality.

\mypara{Performance on Code Translation.}
This task demands both semantic preservation and strict functional correctness, where the limitations of serialization become particularly evident. The +GraphText baseline performs poorly, with a substantial drop in \textit{Execution Accuracy}, suggesting that linearized graphs fail to convey essential program logic. Although both +LoRA and our +\method achieve strong \textit{CodeBLEU} scores, our approach consistently outperforms on fine-grained metrics such as \textit{Syntax Match}, \textit{Dataflow Match}, and \textit{Execution Accuracy} (\emph{e.g.}, 98.26\% on Qwen2.5-Coder-7B). These findings underscore the importance of a dedicated integration mechanism—such as our Bridge module—for effectively translating structural signals into functionally correct code.

\subsection{Robustness to Variable Renaming }
To test whether our framework alleviates the structural information loss in text-only models, we perform a robustness evaluation under code obfuscation. 
The code summarization test set is modified by replacing all function names and internal variables with random identifiers (\emph{e.g.}, \texttt{get\_price} → \texttt{fn\_abc}, \texttt{cost} → \texttt{var\_xyz}), while preserving external library calls \cite{DollmunderstandProgconcepts-counterfactual}. 
As this transformation invalidates lexical metrics such as \textit{ROUGE}, we adopt the semantic-oriented \textit{LLM-as-a-Judge} (LLM-J) score. Results in \autoref{tab:variable_renaming} show pronounced degradation in text-only models, particularly in LoRA variants (\emph{e.g.}, a 15.2\% drop in the 7B LoRA model). In contrast, our +\method remains highly stable, with changes of only -1.6\% and -0.3\% for the 1.5B and 7B models. These findings demonstrate that by leveraging explicit structural graphs, our framework captures true program logic and data flow, effectively resisting superficial identifier variations—a key weakness of text-only approaches.

\input{tab/robust_to_variable_renaming}


\subsection{Ablation Study}
To validate the contribution of each key component within our framework, we conduct two ablation studies on the code translation task using the Qwen2.5-Coder-1.5B.

\begin{figure}[h!]
    \centering
    \includegraphics[width=1\columnwidth]{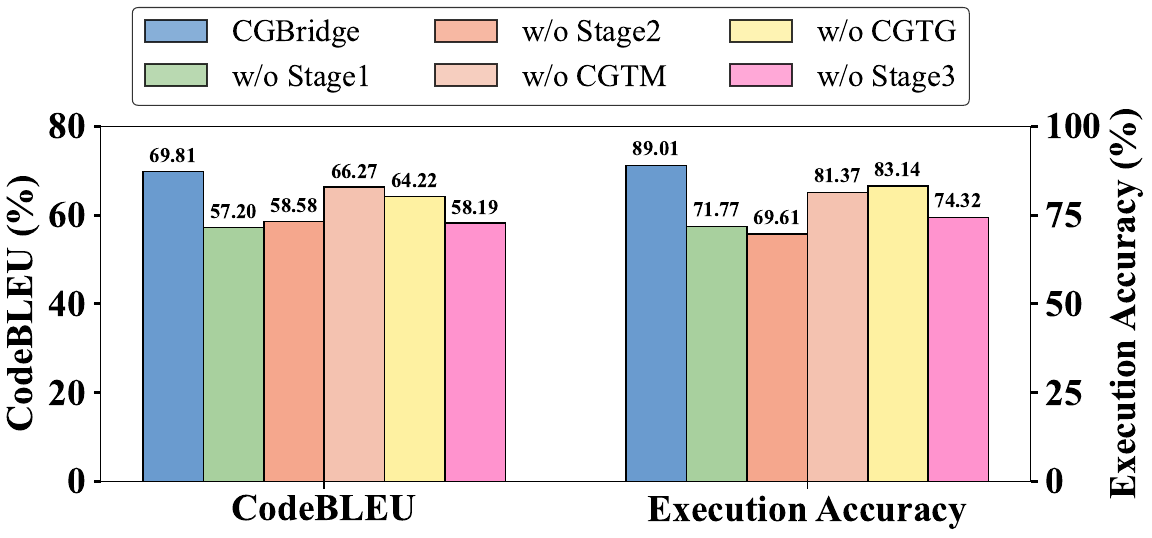}
    \vspace{-2em}
    \caption{Ablation experiments of training components.}
    \label{fig:ablation_training}
    \vspace{-1em}
\end{figure}

\mypara{Impact of Training Components.}
We evaluate the effectiveness of our training strategies by comparing the full \method model with five ablated variants: \textbf{w/o Stage 1 (CGE Pre-training)}, which skips self-supervised pre-training and randomly initializes the CGE module; \textbf{w/o Stage 2 (Cross-Modal Alignment)}, which omits the entire cross-modal alignment and initialized Bridge module with pre-trained BERT weights for the final adaptation; \textbf{w/o GTM} and \textbf{w/o GTG}, which remove their respective alignment objectives from Stage 2; and \textbf{w/o Stage 3 (Task Adaptation)}, which performs only Stage 1 and 2 without the final instruction-based adaptation. 
As shown in \autoref{fig:ablation_training}, removing any component degrades performance, confirming that each part is essential. 
A strong initial graph representation is critical, as skipping Stage 1 results in a sharp drop (CB: 69.81$\to$57.20, EA: 80.91$\to$71.77). 
Removing Stage 2 yields the worst performance (69.61\% EA), while removing Stage 3 also causes a notable decline to 74.32\% EA, demonstrating the importance of both cross-modal alignment and task-specific tuning. Within Stage 2, eliminating GTM or GTG objectives reduces performance on both metrics, highlighting their complementary roles in establishing a robust and precise bridge between modalities.

\mypara{Impact of Graph Components.} 
We evaluate our heterogeneous graph design by comparing it to variants incorporating different structural configurations (Table~\ref{tab:ablation_graph}). The results highlight the synergistic benefits of integrating both syntactic and semantic structures. A model leveraging only the syntactic information from the AST forms a strong baseline, achieving an impressive 84.04\% Execution Accuracy (EA). Meanwhile, a variant trained exclusively on semantic graphs (CFG+DFG) performs competitively, attaining the highest structural similarity score of 75.33\%. However, the full model—combining the syntactic backbone of the AST with the semantic depth provided by CFG and DFG (AST+CFG+DFG)—achieves the best overall performance, with an EA of 89.11\%. These findings suggest that while each structural view contributes valuable information independently, their combination is essential for capturing the comprehensive functional semantics necessary for optimal correctness.
\input{tab/ablation_graph}

\subsection{Efficiency of \method}
\label{efficiency_analysis}
As motivated in our introduction, a primary limitation of many existing LLM adaptation methods is the high computational cost. 
\autoref{tab:efficiency} provides a efficiency analysis of our \method compared with +LoRA and +GraphText baselines.
To begin with, \textbf{\method demonstrates remarkable inference efficiency}. Despite having more trainable parameters, it achieves over 4$\times$ faster inference than LoRA on both 1.5B and 7B models, with negligible latency compared to the vanilla LLM. This advantage arises from a single upfront graph-based computation, whereas LoRA incurs cumulative overhead at every layer during auto-regressive decoding, making \method more suitable for latency-sensitive applications. Furthermore, \textbf{\method exhibits excellent scalability in parameter efficiency}: its fixed parameter size (180.80M) remains constant regardless of LLM scale. Although this count exceeds LoRA (18.46M for 1.5B and 43.12M for 7B), the ratio decreases sharply with model growth—from $\sim$11.5\% (1.5B) to $\sim$2.5\% (7B), and a projected $\sim$0.26\% at 70B—ensuring superior scalability and making \method a future-proof solution for increasingly large foundation models.

\input{tab/efficiency}

\subsection{Performance \textit{v.s.} Code Complexity}
\label{sec:performance_vs_length} 

\begin{figure}[h]
    \centering
    \includegraphics[width=1\columnwidth]{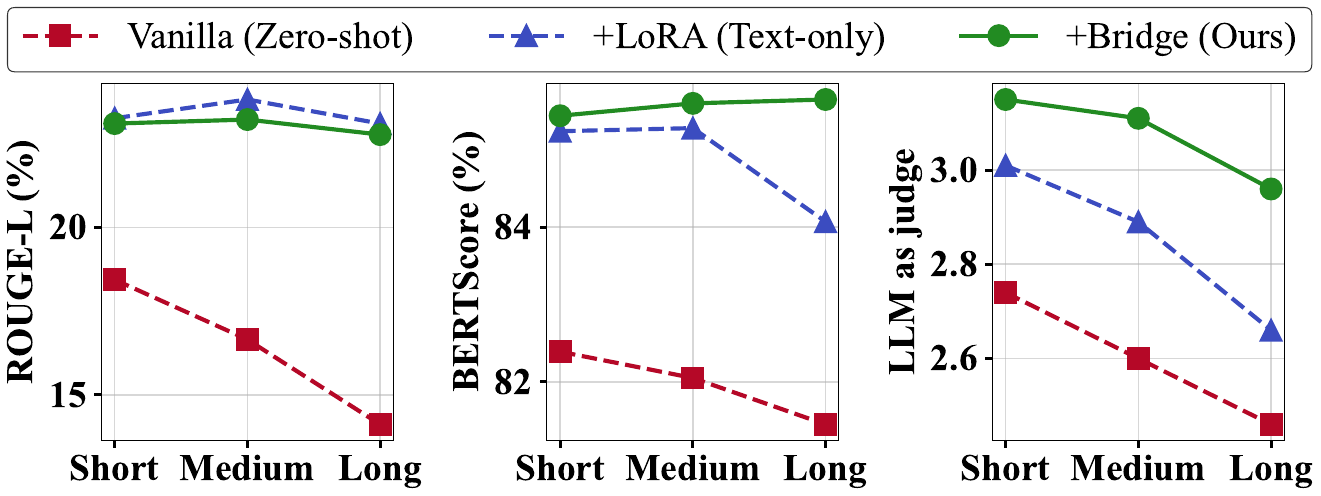}
    \vspace{-2em}
    \caption{Performance across different code length bins.}
    \label{fig:performance_vs_length}
    \vspace{-1em}
\end{figure}

To assess robustness, we evaluate performance on the code summarization task across three code length bins (Short, Medium, Long), using length as a proxy for complexity (\autoref{fig:performance_vs_length}). 
While all methods degrade on longer code, the drop is much smaller for \method, especially on semantic-aware metrics (\textit{BERTScore} and \textit{LLM-as-a-Judge}).
As code length increases, LoRA suffers sharp performance decline (notably in \textit{LLM-as-a-Judge}), whereas our method maintains a more stable and higher level. This highlights the benefit of the structural scaffold from code graphs, which helps the LLM manage long-range dependencies and complex logic-a known limitation for text-only models.


\subsection{Hyper-parameter Analysis}
To validate our design and assess the sensitivity of \method, we study two key hyperparameters: the number of GNN layers $L_g$ in the CGE and the number of Bridge queries $N_q$. Experiments are conducted on the code translation task (Qwen2.5-Coder-1.5B), using \textit{Execution Accuracy} (EA) as the primary metric.

\begin{figure}[!t]
    \centering
    \includegraphics[width=1\columnwidth]{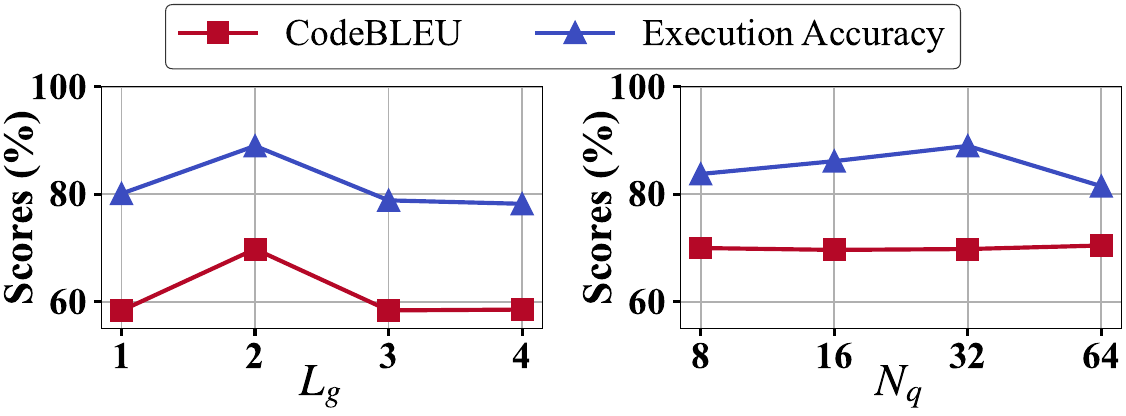}
    \vspace{-2em}
    \caption{Analysis of the hyper-parameter $N_q$ and $L$.}
    \label{fig:hyper-para}
    \vspace{-1em}
\end{figure}

\mypara{Analysis on the $L_g$.}
$L_g$ is crucial for capturing multi-hop dependencies. 
\autoref{fig:hyper-para} (left) shows that performance peaks at $L_g=2$. A single-layer GNN lacks capacity for capturing structural patterns, while deeper GNNs ($L_g>2$) may suffer from over-smoothing \cite{deeperinsightintogcn-oversmoothing} and redundancy in the heterogeneous code graph, as CFG and DFG edges already encode many shortcuts.


\mypara{Analysis on the $N_q$.} 
$N_q$ governs the information capacity of the Bridge module in extracting structural knowledge.
As illustrated in \autoref{fig:hyper-para} (right), \textit{Execution Accuracy} peaks at $N_q = 32$, while \textit{CodeBLEU} remains relatively stable across different values.
This indicates that $N_q = 32$ achieves a favourable trade-off, providing sufficient representation power while mitigating the noise introduced by excessive queries.
A small $N_q$ may limit information flow, whereas a large $N_q$ risks signal dilution.


\begin{figure}[!h]
    \centering
    \vspace{-2mm}
    \includegraphics[width=.99\columnwidth]{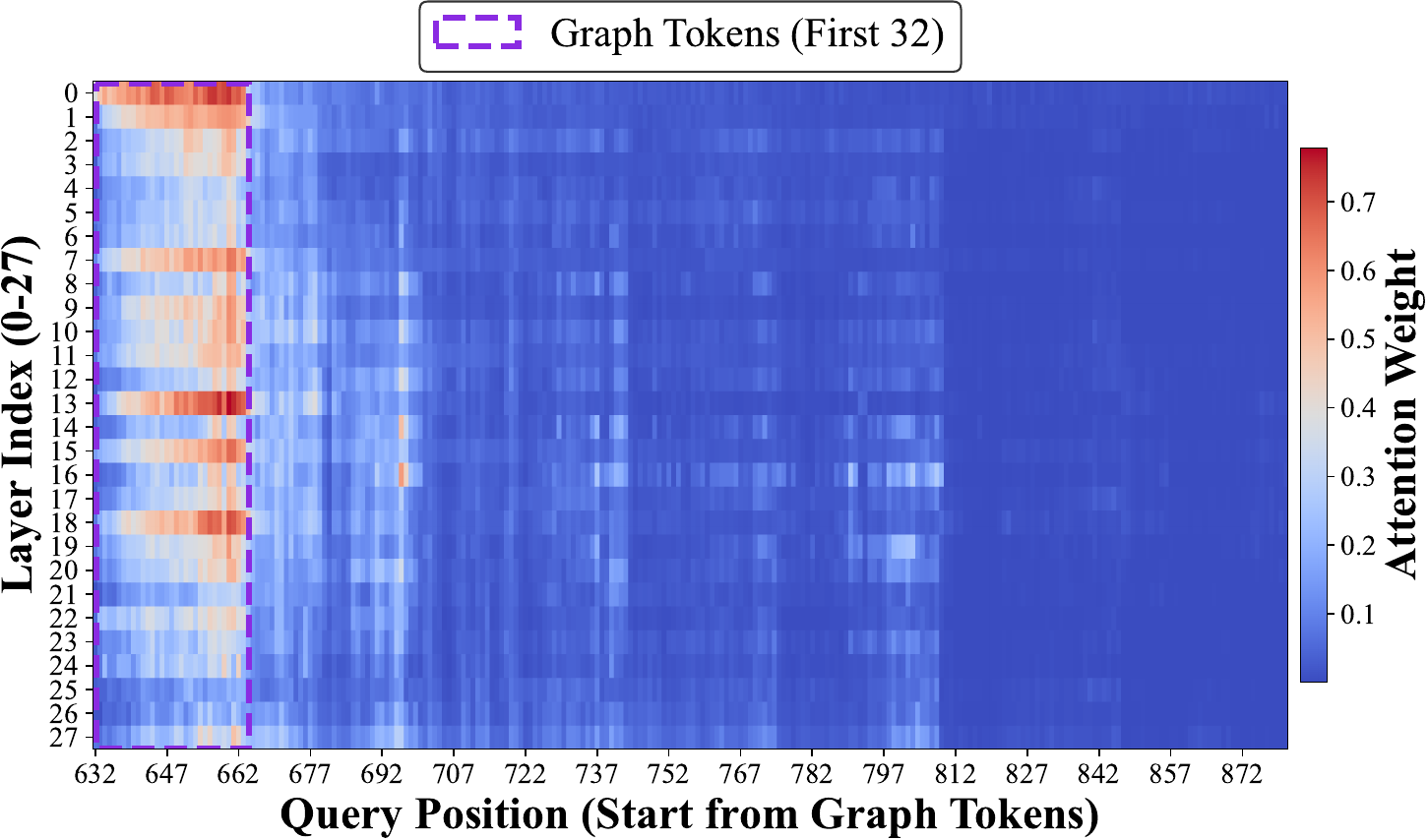}
    \vspace{-1em}
    \caption{Static attention patterns at the prompting stage.}
    \label{fig:prompt_attention_heatmap}
    \vspace{-1em}
\end{figure}

\subsection{Qualitative Analysis} 
\label{qualitative_analysis}
To interpret our model’s behavior, we visualize attention patterns on the code summarization task using Qwen2.5-Coder-1.5B. This form of attention-based analysis has been widely adopted in prior studies to understand transformer behavior in language and code tasks~\cite{visiontransformersneedregisters,imageworth16x16words}. We track how the model attends to the 32 graph-derived prompt tokens $P_G$, averaged over heads and samples, to assess interaction with structural cues.
We also present a concrete example to illustrate \method's effectiveness.

\mypara{Static Attention at the Prompt Stage.} 
\autoref{fig:prompt_attention_heatmap} shows that lower-to-mid layers exhibit concentrated attention on graph tokens, suggesting early structural grounding. As layers deepen, attention becomes more diffuse, indicating layer specialization.
Notably, attention on graph tokens remains strong even for late-position text tokens (\emph{e.g.}, >740), showing that the model persistently refers to the graph context as a global ``anchor'' to resolve long-range dependencies.


\begin{figure}[!t]
    \centering
    \includegraphics[width=1\columnwidth]{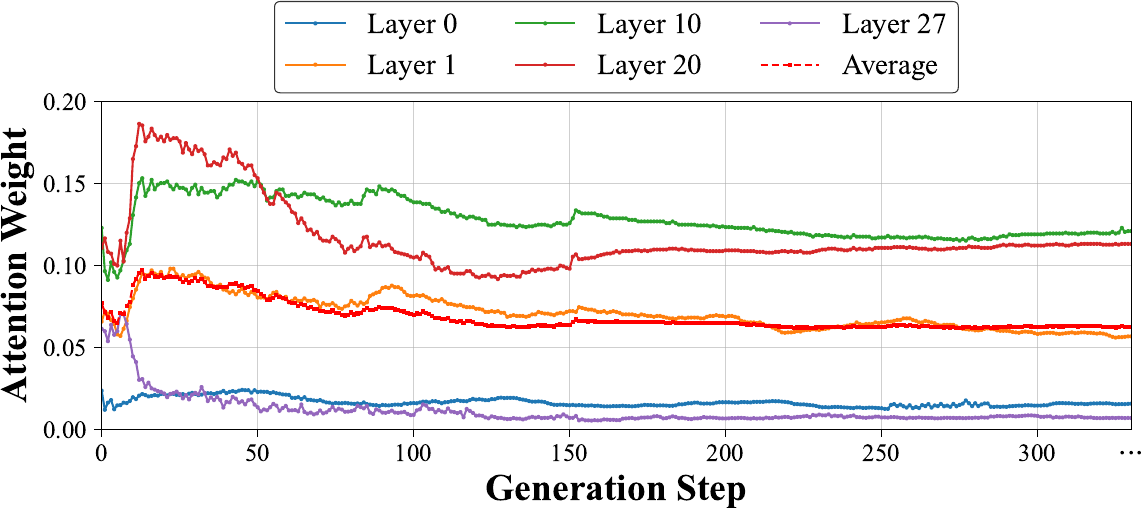}
    \vspace{-2em}
    \caption{Average attention to graph tokens across generation steps, grouped by representative layers.}
    \label{fig:generation_attention_trend}
    \vspace{-1em}
\end{figure}

\mypara{Dynamic Attention during Generation. }
\autoref{fig:generation_attention_trend} shows the average attention from each newly generated token to the 32 graph tokens during decoding. 
Attention remains consistently high across all steps, highlighting the structural anchoring effect of \method. 
Different layers exhibit distinct patterns: middle layers (\emph{e.g.}, Layer 10) show stable attention, serving as persistent structural anchors, while later layers (\emph{e.g.}, Layer 20) display sharp peaks, likely refining or reinterpreting structural cues. This layered and sustained attention suggests that CGBridge enhances decoding robustness, especially for long and complex code.



\begin{figure}[!t]
    \centering
    \includegraphics[width=1\columnwidth]{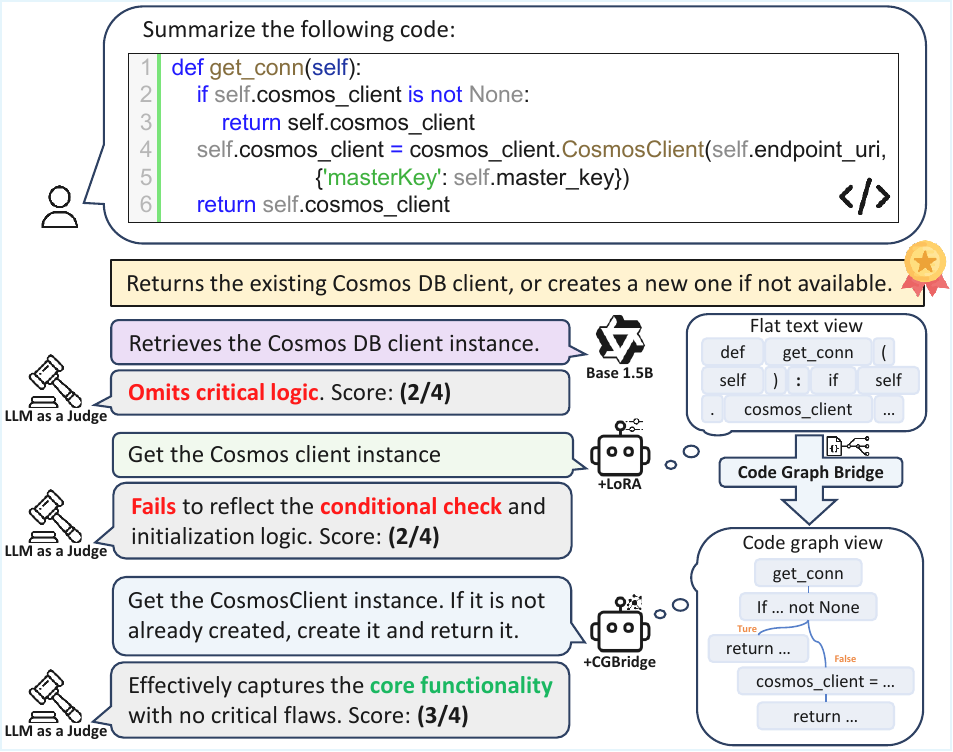}
    \vspace{-2em}
    \caption{Case study.}
    \vspace{-1em} 
    \label{fig:case_study}
\end{figure}

\mypara{Case Study.}
To demonstrate \method's superior logical understanding, we present a case study on a Python function using lazy initialization (\autoref{fig:case_study}). 
Compared to base and +LoRA models, which produce fluent but incomplete summaries missing the critical conditional logic (scored 2/4 by LLM-as-a-Judge), \method leverages structural information from the code graph.
This enables it to correctly interpret the control flow and generate a more accurate summary (scored 3/4). The case clearly shows how CGBridge enhances program comprehension by integrating structural cues beyond flat text.


\section{Related Work}
\label{append: related-work}



\mypara{Code Representation Learning.}
Transforming source code into numerical representations that capture its semantics and structure is a key challenge. Early models treated code as token sequences, applying NLP Transformers to linearized code, achieving strong results on large codebases \cite{CodeBERT,wang2021codet5}. However, this approach suffers from structural information loss \cite{COMEX-structualblindness}, as it discards explicit code structure, limiting its ability to capture deeper semantics and making it vulnerable to adversarial attacks like ``If-Else Flip'' and ``Variable Renaming'' \cite{ContraBERT}. To address this, a second paradigm emerged, representing code as explicit graphs using structures like ASTs \cite{AST-code2seq}, CFGs \cite{CFG-androidapp}, and DFGs \cite{guo2021-graphcodebert}. Later works combined multiple graph types into holistic representations like Code Property Graphs (CPGs) \cite{multi-view-graph-rep-for-prog}, designed for GNNs or traditional program analysis tools. Despite their structural fidelity, this paradigm faces challenges such as architectural mismatches with permutation-invariant GNNs \cite{MISIM} and overhead in graph processing. Recently, hybrid approaches have emerged, integrating structural information into sequence-based models \cite{guo2021-graphcodebert,Graphix-T5}.

\mypara{LLMs for Code.}
Instruction-tuned code LLMs such as CodeLlama \cite{codellama}, Qwen-Coder \cite{qwen2.5coder}, and Deepseek-coder \cite{guo2024deepseekcoder}  have brought significant advances, offering strong zero-shot capabilities on code tasks. These models are often adapted from general-purpose LLMs via continued pretraining on massive code datasets. For downstream adaptation, Parameter-Efficient Fine-Tuning (PEFT) \cite{peft} methods like LoRA \cite{lora} are widely used. However, such sequential fine-tuning is computationally costly and limited by its token-level view of code, often missing structural dependencies.
 
\mypara{Integrating Graph Structures with LLMs.}
Bridging the gap between non-sequential graph structures and sequential LLMs is a key challenge. Approaches to infusing LLMs with structural knowledge can be divided into two main paradigms: graph serialization and hybrid architectures. The first, graph serialization, translates the graph into a linear text format for direct LLM processing, using methods like adjacency lists or natural language descriptions~\cite{zhao2023graphtext,guo2023gpt4graph}. However, this approach often struggles to preserve complex topologies and can result in long input sequences. The second paradigm uses hybrid GNN–LLM architectures, where a GNN encodes structural information to be aligned with the LLM \cite{tang2024graphgpt,he2024gretrieverr}. Additionally, frameworks like CODEXGRAPH~\cite{liu2024codexgraph} and RepoGraph~\cite{ouyang25repograph} treat the graph as an external, queryable resource to enhance LLM performance. However, this method is sensitive to retrieved context, prone to information loss, and limited by the LLM’s context window.


\section{Conclusion}
In this paper, we introduced \method, a novel plug-and-play framework for augmenting LLMs with an external, trainable bridge module that captures rich structural information from code. 
\method leverages a GNN-based encoder pre-trained on a large-scale code graph dataset to learn structural semantics, and employs a bridge module to align representations across code, graph, and text via cross-modal attention. 
This module generates structure-informed prompts, which are injected into a frozen LLM and fine-tuned for downstream tasks.
Our experiments show that \method significantly outperforms text-only and graph-augmented baselines, especially in ensuring functional correctness and robustness to complexity.
Visualization and ablation studies revealed that the graph guides the model to prioritize function over form. Notably, the framework is highly practical, achieving a 4× inference speedup over LoRA. Our work validates that modularly infusing structural knowledge is a promising and efficient paradigm for advancing code intelligence.

\clearpage
\balance
\bibliographystyle{ACM-Reference-Format}
\bibliography{ref/reference}
\appendix
\clearpage

\input{docs/appendix}

\end{document}

%% file: tab/MainResults.tex
\begin{table*}[t!]
\linespread{0.95}\selectfont 
    \centering
    \renewcommand{\arraystretch}{1.1}
    
    
    \caption{Code Summarization (left) and Code Translation (right) results across model groups. Metrics include METEOR, ROUGE-L, SBCS, B-Score, LLM-J, CodeBLEU (CB), Syntax Match (SM), Dataflow Match (DM), and Execution Accuracy (EA). All scores are in \%, except LLM-J (0–4). Bold denotes the best per group.}

    \vspace{-1em}
    \label{tab:MainResults}
    
    \setlength{\tabcolsep}{4pt} 

    \begin{tabular*}{\textwidth}{@{\extracolsep{\fill}} l ccccc ccccc}

        \toprule
        \multirow{2}{*}{Method}& \multicolumn{5}{c}{\textbf{Code Summarization}} & \multicolumn{4}{c}{\textbf{Code Translation}} \\
        
        \cmidrule(lr){2-6} \cmidrule(lr){7-10}
        
         & METEOR & ROUGE-L & SBCS & B-Score& LLM-J & CB & SM& DM& EA\\
        \midrule

        Qwen2.5-Coder-1.5B       & 16.95 & 16.40 & 56.68 & 81.96 & 2.70  & 58.89 & 71.43 & 40.74 & 70.63 \\
        \hspace{1em}+LoRA   & 16.73 &  \textbf{23.40} & 56.97 & 84.86 & 2.85  & 63.76 & 76.64 & 63.44 & 84.68 \\
        \hspace{1em}+GraphText & 15.93 &  13.69 & 51.61 & 81.36 & 2.66  & 55.68 & 64.27 & 54.28 & 57.26 \\
        \hspace{1em}+\method   & \textbf{19.76} &  23.04   & \textbf{59.12} & \textbf{85.56} & \textbf{3.18} & \textbf{69.81} & \textbf{78.80} & \textbf{68.55} & \textbf{89.01} \\
        \hline
        
        Qwen2.5-Coder-7B        & 18.31 & 19.91 & 59.13 & 83.57 & 2.78  & 63.39 & 74.67 & 48.07 & 89.46 \\
        \hspace{1em}+LoRA   & \textbf{20.94} & \textbf{23.50} & 60.02 & 86.02 & 2.83  & 62.88 & 77.70 & 65.03 & 97.17 \\
        \hspace{1em}+GraphText & 18.05 &  15.94 & 53.18 & 82.77 & 2.96  & 63.52 & 67.85 & 67.52 & 70.75 \\
        \hspace{1em}+\method & 15.73 & 20.08 & \textbf{61.62} & \textbf{86.61} & \textbf{3.23} & \textbf{73.91} & \textbf{77.20} & \textbf{74.18} & \textbf{98.26} \\
        \hline

        CodeLlama-7B            & 17.51 & 18.71 & 55.85 & 82.91 & 2.83 & 27.64 & 54.65 & 46.05 & 71.76 \\
        \hspace{1em}+LoRA & \textbf{21.83} & \textbf{25.31} & 60.44 & 86.23 & 2.89 & 69.10 & 74.14 & 70.88 & 93.17 \\
        \hspace{1em}+GraphText & 15.00 &  14.30 & 49.50 & 81.60 & 2.87  & 57.81 & 62.84 & 59.02 & 60.32 \\
        \hspace{1em}+\method & 18.70 & 23.38 & \textbf{60.65} & \textbf{88.35} & \textbf{2.99} & \textbf{71.32} & \textbf{79.21} & \textbf{74.49} & \textbf{95.07} \\
        \hline
        
        Deepseek-coder-1.3B  & 10.79 & 6.12 & 45.30 & 79.08 & 1.63 & 48.82 & 67.67 & 55.58 & 51.37 \\
        \hspace{1em}+LoRA & 11.67 & 5.88 & 49.74 & 80.05 & 1.87 & 52.03 & \textbf{69.73} & 54.31 & 63.06 \\
        \hspace{1em}+GraphText  & \textbf{15.45} & 14.76 & 48.76 & 81.86 & 1.94  & \textbf{55.02} & 60.39 & 54.22 & 69.31 \\
        \hspace{1em}+\method  & 13.42 & \textbf{16.30} & \textbf{50.44} & \textbf{84.02} & \textbf{2.61}  & 50.07 & 62.22 & \textbf{55.44} & \textbf{74.49 } \\

        \hline
        Deepseek-coder-6.7B  & 19.28 & 17.33 & 56.79 & 82.29 &2.93  & 61.34 & 73.21 & 56.20 & 90.93 \\
        \hspace{1em}+LoRA & 19.58 & 21.11 & 58.21 & 84.30 & 2.97 & \textbf{64.40} & 73.12 & \textbf{70.67} & 93.88 \\
        \hspace{1em}+GraphText & 16.46&  16.90 & 49.62& 82.64 & 2.72  & 58.42 & 62.61 & 60.01 & 53.97 \\
        \hspace{1em}+\method  & \textbf{19.96} & \textbf{24.79} & \textbf{59.10} & \textbf{85.75} &\textbf{3.22} & 64.32 & \textbf{73.27} & 67.79 & \textbf{95.92} \\

        \bottomrule
    \end{tabular*}
    \vspace{-1em}
\end{table*}

%% file: tab/robust_to_variable_renaming.tex
\begin{table}[t!]
\linespread{0.95}\selectfont 

    \centering
    \caption{Robustness to variable renaming.}
    \vspace{-1em}
    \label{tab:variable_renaming}

    \begin{tabularx}{\columnwidth}[c]{l *{3}{>{\centering\arraybackslash}X}}
        \toprule
        \textbf{Method} & \textbf{Original} & \textbf{Obfuscated} & \textbf{Change} \\
        \midrule

        Qwen2.5-Coder-1.5B & 2.70 & 2.59 & -4.1\% \\
        \hspace{1em}+LoRA & 2.85 & 2.36 & -17.2\% \\
        \hspace{1em}+GraphText & 2.66 & 1.90 & -28.6\% \\
        \hspace{1em}+\textbf{\method}  & 3.18 & 3.13 & \textbf{-1.6\%} \\

        \midrule

        Qwen2.5-Coder-7B & 2.78 & 2.71 & -2.5\% \\
        \hspace{1em}+LoRA  & 2.83 & 2.40 & -15.2\% \\
        \hspace{1em}+GraphText  & 2.96 & 2.13 & -28.4\% \\
        \hspace{1em}+\textbf{\method} & 3.23 & 3.22 & \textbf{-0.3\%} \\

        \bottomrule
    \end{tabularx}
    \vspace{-4mm} 
\end{table}

%% file: tab/ablation_graph.tex
\begin{table}[!t]
\linespread{0.95}\selectfont 

    \centering
    \caption{Ablation experiments of graph components.}
    \setlength{\tabcolsep}{8pt} 
    \label{tab:ablation_graph}
    \vspace{-1em}
        \begin{tabular}{@{}lcccc@{}} 
        \toprule
        \textbf{Graph Components} & \textbf{CB}  & \textbf{SM} & \textbf{DM} &  \textbf{EA} \\
        \midrule
        AST & 65.99 & 77.26 & 51.64 & 84.04 \\ 
        AST + CFG & 74.90 & \textbf{79.81} & 66.42  & 85.28 \\ 
        AST + DFG & 73.46 & 79.10 & 69.54 & 83.23 \\ 
        CFG + DFG & \textbf{75.33} & 79.52 & \textbf{70.57} & 88.54 \\ 
        
        AST + CFG + DFG & 69.81 & 78.80 & 68.55 & \textbf{89.11} \\
        
        \bottomrule
    \end{tabular}
    \vspace{-2mm}
\end{table}

%% file: tab/efficiency.tex
\begin{table}[t]
\linespread{0.95}\selectfont
    \centering

    \caption{Efficiency analysis.}
    \vspace{-4mm}
    \label{tab:efficiency}

    \begin{tabularx}{\columnwidth}[c]{l *{3}{>{\centering\arraybackslash}X}}
        \toprule
        
        \textbf{Method} & 
        \makecell{\textbf{Trainable}\\\textbf{Params}} & 
        \makecell{\textbf{Training}\\\textbf{Time}\textsuperscript{a}}  & 
        \makecell{\textbf{Inference}\\\textbf{Latency}\textsuperscript{b}} \\
        
        \midrule 

        Qwen2.5-Coder-1.5B & - & - & $\sim$202 ms \\
        \hspace{1em}+LoRA & 18.46 M & 56.2 s & $\sim$969 ms \\
        \hspace{1em}+GraphText & - & - & $\sim$598 ms \\
        \hspace{1em}+\textbf{\method} & 180.80 M & 67.5 s & $\sim$215 ms \\

        \midrule

        Qwen2.5-Coder-7B & - & - & $\sim$299 ms \\
        \hspace{1em}+LoRA & 43.12 M & 110.5 s & $\sim$1463 ms \\
        \hspace{1em}+GraphText & - & - & $\sim$1197 ms \\
        \hspace{1em}+\textbf{\method} & 180.80 M & 272.1 s & $\sim$371 ms \\

        \bottomrule
    \end{tabularx}
    
    \footnotesize 
    \raggedright 
    \textsuperscript{a} Measured per epoch. 
    \textsuperscript{b} Measured per sample. 
    \vspace{-1em}
\end{table}

%% file: docs/appendix.tex
\section{Details of the Code Graph Dataset}
\label{sec:detail_of_codegraphdataset}

\subsection{Detailed Node and Edge Type Attributes}
\label{sec:nodes_edges_textual_attributes}
\input{tab/textual_attributes}

\subsection{Graph Statistics of Datasets}
\autoref{tab:stat_graph} presents the structural statistics of our curated datasets. The significant number of nodes and edges in an average code graph highlights the structural complexity that text-based models often overlook. This data underscores the necessity of a structure-aware approach like \method.

\begin{table}[h!]
\centering
\caption{Average structural statistics for the curated code graph datasets.}
\label{tab:stat_graph}
\begin{tabular}{lcc}
\toprule
\textbf{Statistic} & \textbf{CodeSearchNet} & \textbf{XLCoST} \\
\midrule
Total Samples & 261,372 & 10,344 \\
Avg. Nodes & 125.41 & 145.77 \\
Avg. AST Edges & 124.41 & 144.77 \\
Avg. CFG Edges & 16.21 & 27.95 \\
Avg. DFG Edges & 22.85 & 32.23 \\
\bottomrule
\end{tabular}
\end{table}

\section{Detailed InfoNCE Loss Formulation}
\label{sec:infonce}

This section provides the detailed mathematical formulation for the symmetric InfoNCE loss used in \autoref{eq:gtc_loss} for Graph-Text Contrastive Learning (GTC). For a given batch of $M$ graph-text pairs, the loss is fully expanded as:
\begin{equation} \label{eq:gtc_loss_full}
\begin{split}
\mathcal{L}_{\text{GTC}} = -\frac{1}{2M} \sum_{i=1}^{M} \bigg( & \log \frac{\exp(s(B_Q^i, h_{\text{cls}}^i)/\tau)}{\sum_{j=1}^{M} \exp(s(B_Q^i, h_{\text{cls}}^j)/\tau)} \\
& + \log \frac{\exp(s(h_{\text{cls}}^i, B_Q^i)/\tau)}{\sum_{j=1}^{M} \exp(s(h_{\text{cls}}^i, B_Q^j)/\tau)} \bigg) \,,
\end{split}
\end{equation}
where:
\begin{itemize}[leftmargin=*,nosep]
    \item $B_Q^i$ is the graph embedding for the $i$-th sample in the batch.
    \item $h_{\text{cls}}^i$ is the text embedding for the $i$-th sample in the batch.
    \item $s(u, v)$ is the similarity function. In our implementation, it is the maximum cosine similarity computed over all N\textsubscript{q} query tokens.
    \item $\tau$ is the learnable temperature parameter that scales the similarity scores.
\end{itemize}

\section{Implementation of GNN layer}
\label{sec:CodeGraphEncoder}
The Code Graph Encoder, denoted as $\text{CGE}(\cdot)$, is responsible for transforming the input code graph $G$ into a graph-level embedding $x_G = \text{CGE}(G)$. 
This embedding encapsulates the essential structural and semantic characteristics of the code.

The CGE module is implemented as a $L_g$ layers graph Transformer~\cite{yunsheng-graphtransformerconv}. 
Let $h_i^{(l-1)}$ be the feature vector of node $i$ at the input of layer $l$ (with $h_i^{(0)}$ being the initial node features from $\mathcal{X}_\mathcal{V}$) and let $e_{ji} \in \mathcal{X}_\mathcal{E}$ be the feature vector for the edge connecting node $j$ to node $i$.
This update involves an attention mechanism where the coefficient $\alpha_{ji}^{(l)}$, representing the attention node $i$ pays to a neighbor $j$, is computed as:
\begin{equation}
\alpha_{ji}^{(l)} = \text{softmax}_j \left( \frac{(W_Q^{(l)}h_i^{(l-1)})^{\top} (W_K^{(l)}h_j^{(l-1)} + W_E^{(l)} e_{ji})}{\sqrt{d_k}} \right).
\end{equation}
Here, $W_Q^{(l)}$, $W_K^{(l)}$, and $W_E^{(l)}$ are learnable weight matrices for projecting the query (node $i$), key (node $j$), and edge feature ($e_{ji}$) components at layer $l$, respectively, while $d_k$ is the key dimensionality. 
The term $W_E^{(l)} e_{ji}$ denotes the linearly transformed edge feature, which modulates the attention scores during key computation.

The aggregated representation $\tilde{h}_i^{(l)}$ for node $i$ is then formed by combining its self-representation with the attention-weighted messages from its neighbors:
\begin{equation}
    \tilde{h}_i^{(l)} = W_\text{self}^{(l)} h_i^{(l-1)} + \sum_{j \in \mathcal{N}(i)} \alpha_{ji}^{(l)} \left( W_\text{val}^{(l)} h_{j}^{(l-1)} + W_E^{(l)} e_{ji} \right),
\end{equation}
where $W_{\text{self}}^{(l)}$ and $w_{\text{val}}^{(l)}$ are learnable weight matrices. 
The edge features $e_{j i}$, transformed by the same matrix $W_E^{(l)}$ as in the attention mechanism, also contribute to the aggregated message from each neighbor.

For intermediate layers, these aggregated representations $\tilde{H}^{(l)} = \{ \tilde{h}_{v_i}^{(l)} \mid \forall v_i \in \mathcal{V} \}$ are processed through a sequence of batch normalization, ReLU activation, and dropout to yield $H^{(l)}$. 
The graph-level representation $x_G \in \mathbb{R}^{d_{out}}$, which serves as the input from the graph modality to the Bridge module, is processed with a readout in the last layer for final node representation.

\section{Prompt Templates}
\label{sec:prompt}
This section provides the exact instruction templates used during the instruction-based task adaptation and for LLM-as-a-judge evaluation. 

\subsection{Code summarization}
The goal is to generate a high-quality docstring for a given code snippet. The prompts are designed to enforce a strict format consistent with the PEP 257 standards.

\begin{promptbox}
    \textbf{System Prompt}: 
    You are an expert Python assistant. Generate clear, concise, and accurate docstrings strictly following PEP 257 triple quote formatting.

    \par\vspace{2ex}
    
    \textbf{User Prompt}: 
    Generate a Python docstring for the code below. \\
    Format: PEP 257 triple quotes.\\
    Required Structure:
    \begin{itemize}[leftmargin=*,nosep]
        \itemsep0em  
        \item[-] One-line summary of the function. (Optional) More detailed explanations are provided if the logic is complex.
        \item[-] Parameters: param\_name (param\_type): Description of parameter. (Use 'None' if no parameters)
        \item[-] Returns: return\_type: Description of returned value. (Use 'None' if no explicit return)
    \end{itemize}

    Code:
    
    \codemore{python
      
      \{code\}
      
      }
      
\end{promptbox}

\subsection{Code Translation}
The goal is to translate a function from a source programming language to a target language, ensuring functional equivalence and producing runnable code.

\begin{promptbox}
    \textbf{System Prompt}:
    
    Role: Python to Java Translator.      
    
    Output: Single, runnable Java class (e.g., \code{Solution} or \code{Main}).
    
    
    Structure:
    \begin{itemize}[leftmargin=*,nosep]
        \itemsep0em 
        \item[1.] Core logic in a  \code{public static} method.
        \item[2.]  \code{public static void main(String[] args)}: Calls static method (use user's sample inputs if given, else defaults) and prints output via \code{System.out.println()}.
    \end{itemize}
    
    Ensure:
    \begin{itemize}[leftmargin=*,nosep]
        \itemsep0em 
        \item[-] Functional Python-Java equivalence.
        \item[-] Correct Java types and standard library mapping.
        \item[-] Necessary \code{import}s.
        \item[-] Compilable and runnable code.
    \end{itemize}

    \par\vspace{1ex}
    
    \textbf{User Prompt}:
    
    Translate the following Python code to Java.
    
      \codemore{python
      
      \{code\}
      
      }
    
\end{promptbox}

\subsection{LLM as a Judge}
For a reference-free evaluation of the Code Summarization task, we employ a powerful LLM (e.g., GPT-4) as an evaluator. The following prompts guide the LLM to score the generated summaries based on predefined criteria, with a heavy emphasis on consistency with the source code.

\begin{promptbox}
    \textbf{System Prompt}:

    You are an expert Principal Software Engineer acting as a meticulous Code Reviewer. Your sole task is to provide a critical and objective evaluation of a candidate code summary based on the provided source code.

    Your evaluation must follow these steps:
    \begin{enumerate}[leftmargin=*,nosep]
        \itemsep0em 
        \item Carefully read the source code to fully understand its functionality, inputs, outputs, and key logic.
        \item Critically analyze the candidate summary against the code.
        \item Provide a structured evaluation based on the four dimensions below.
    \end{enumerate}

    \textbf{Evaluation Dimensions}:
    \begin{itemize}[leftmargin=*,nosep]
        \itemsep0em 
        \item[-]  \textbf{Coherence (0-4)}: How logically organized and well-structured is the summary? Does it form a coherent description of the code? (0=Incoherent, 4=Perfectly coherent).
        \item[-] \textbf{Consistency (0-4)}: Does the summary accurately reflect the code's functionality and logic? Are there any factual errors or hallucinations? This is the most critical dimension. (0=Contradicts the code, 4=Perfectly consistent).
        \item[-] \textbf{Fluency (0-4)}: Is the summary written in clear, natural, and grammatically correct language? (0=Unreadable, 4=Perfectly fluent).
        \item[-] \textbf{Relevance (0-4)}: Does the summary capture the essential information without including redundant or trivial details? (0=Irrelevant, 4=Perfectly relevant).
    \end{itemize}

    \textbf{Overall Score}:

    After rating the four dimensions, provide a holistic \textbf{Overall Score (0-4)}. This score is NOT a simple average. You must weigh \textbf{Consistency} most heavily, as an inconsistent summary is fundamentally flawed, regardless of its fluency.
    
    \par\vspace{2ex}

    \textbf{User Prompt}:

    Please evaluate the following code summary.

    \par\vspace{1ex}
    
    \textbf{Source Code}:

    \codemore{python
    
    \{code\}
    
    }

    \textbf{Candidate Summary}:
    
    \{candidate\}
    
\end{promptbox}

\section{Implementation Details of Experiments.} 
\label{sec:implementation_details}
We implement the CGE using a 2-layer Graph Transformer, with the Bridge Module utilizing $N_q=32$ query tokens. The Bridge Module is initialized with the weights of a pre-trained BERT-base-uncased model \cite{devlin-bert}, adopting its original architecture, including the number of layers $L_b$. The initial node and edge features are encoded using UniXcoder-base \cite{guo2021-unixcoder}.
For CGE pre-training and the alignment training in stage 1, we use only the source code of the  Python training split of \textit{CodeSearchNet}. In the adaptation training in stage 2, we use the respective training splits from \textit{CodeSearchNet} (for summarization) and \textit{XLCoST} (for translation).
We conduct experiments on a variety of open-source Code LLMs, including \textit{Qwen2.5-Coder-Instruct} (1.5B, 7B) \cite{qwen2.5coder}, \textit{CodeLlama7b-Instruct-hf} \cite{codellama}, and \textit{Deepseek-coder-Instruct} (1.3B, 6.7B) \cite{guo2024deepseekcoder}.
All experiments are performed on a server equipped with 8 NVIDIA H800 80GB GPUs.

\begin{table*}[h!]
\centering
\caption{Ablation study on the GNN backbone for the Code Graph Encoder (CGE), evaluated on the Qwen2.5-Coder-1.5B model.}
\label{tab:gnn_backbon}
\begin{tabular}{l ccccc ccccc}
\toprule
\multirow{2}{*}{\textbf{GNN Type}} & \multicolumn{5}{c}{\textbf{Code Summarization}} & \multicolumn{5}{c}{\textbf{Code Translation}} \\
\cmidrule(lr){2-6} \cmidrule(lr){7-11}
 & METEOR & ROUGE-L & SBERT-CS & B-Score & LLM-J & CB & BLEU & SM & DM & EA (\%) \\
\midrule
GCN & 19.24 & 22.86 & 57.49 & 85.03 & 3.16 & 64.32 & 62.19 & 72.49 & 56.51 & 81.89 \\
GAT & 17.31 & 21.77 & 54.82 & 84.50 & 2.98 & \textbf{70.11} & 67.67 & 74.03 & \textbf{68.99} & 81.75 \\
Graph Transformer  & \textbf{19.76} & \textbf{23.04} & \textbf{59.12} & \textbf{85.56} & \textbf{3.18} & 69.81 & \textbf{67.75} & \textbf{78.80} & 68.55 & \textbf{89.01} \\
\bottomrule
\end{tabular}
\end{table*}

\subsection{The Choice of Graph Encoder.}
To validate our choice of using a Graph Transformer as the backbone for the Code Graph Encoder (CGE), we conducted an ablation study comparing it against two other common GNN architectures: Graph Convolutional Network (GCN) and Graph Attention Network (GAT). All experiments were performed using the Qwen2.5-Coder-1.5B model as the base LLM. The results for both code translation and code summarization tasks are presented in \autoref{tab:gnn_backbon}. The findings show that the Graph Transformer consistently achieves superior or highly competitive performance across all key metrics, particularly in \textit{Execution Accuracy }for translation and semantic similarity metrics for summarization, thus justifying its selection for our \method framework.

\section{Hyperparameters}
\label{sec:hyperparameters}
Our main experiment hyperparameter configurations are detailed in the tables below, structured by our three-stage training: \autoref{tab:hyperparams_cge} (Stage 1: CGE Pre-training), \autoref{tab:hyperparams_stage2} (Stage 2: Cross-Modal Alignment), and \autoref{tab:hyperparams_stage3} (Stage 3: Instruction-Based Task Adaptation). All experiments, including ablation studies, adhere to these settings unless stated otherwise.


\begin{table*}[h]
\centering
\caption{Hyperparameters for Stage 1: CGE Pre-training.}
\label{tab:hyperparams_cge}
\vspace{-1em}
\begin{tabular}{lc} 
\toprule
\textbf{Parameter} & \textbf{Value} \\ 
\midrule
GNN Type & Graph Transformer \\
Input Dimension & 768 \\
Hidden Dimension & 1024 \\
Output Dimension & 768 \\
GNN Layers $L_g$ & 2 \\
Attention Heads & 4 \\
Dropout Rate & 0.1 \\
Node Feature Drop Rate & 0.05 \\ 
Edge Drop Rate & 0.05 \\
Batch Size & 128 \\ 
Learning Rate & 1e-5 \\
Weight Decay & 0.01 \\
Contrastive Temperature  & 0.3 \\ 
Contrastive Loss Weight $\lambda_{cl}$& 0.6 \\
Edge Prediction Loss Weight $\lambda_{edge}$ & 0.4 \\
Negative Sampling Ratio & 0.5 \\
Early Stopping Patience & 20 \\ 
\bottomrule
\end{tabular}
\end{table*}
  
\begin{table*}[h]
\centering
\caption{Hyperparameters for Stage 2: Cross-Modal Alignment.}
\vspace{-1em}
\label{tab:hyperparams_stage2}
\begin{tabular}{lc} 
\toprule
\textbf{Parameter} & \textbf{Value} \\  
\midrule
Initial Weight & BERT-base-uncased \\  
Bridge layers $L_b$ & 12 \\
Query Tokens $N_q$ & 32 \\
Graph Embedding Dim & 768 \\
Model Dim $d_\text{model}$ & 768 \\
Cross-Attention Frequency & 2 \\
Batch Size & 16 \\  
Contrastive Temperature $\tau$ & 0.07 \\
Learning Rate & 5e-7 \\ 
Weight Decay & 0.01 \\
Warmup Ratio & 0.01 \\
Max Epochs & 200 \\
Scheduler Type & Plateau \\
Early Stopping Patience & 20 \\
\bottomrule
\end{tabular}
\end{table*}

\begin{table*}[h]
\centering
\caption{Hyperparameters for Stage 3: Instruction-Based Task Adaptation.}
\vspace{-1em}
\label{tab:hyperparams_stage3}
\begin{tabular}{lc}
\toprule
\textbf{Parameter} & \textbf{Value} \\
\midrule
Bridge Module & Initialized from Stage 2 \\
Precision & bfloat16 \\
Batch Size & 8 \\
Learning Rate & 1e-6 \\
Weight Decay & 0.01 \\
Max Epochs & 200 \\
Warmup Ratio & 0.01 \\
Early Stopping Patience & 20 \\
Scheduler Type & Plateau \\
Scheduler Patience & 2 \\
Scheduler Factor & 0.5 \\
Min Learning Rate & 1e-10 \\
Temperature & 0 \\
Repetition Penalty & 1.1 \\
\bottomrule
\end{tabular}
\end{table*}

%% file: tab/textual_attributes.tex
\begin{table}[h!]
    \centering
    \caption{Table of textual attributes and their classifications.}
    \vspace{-1em}
    \begin{tabularx}{\columnwidth}{c >{\raggedright\arraybackslash}X}

        \toprule
        \textbf{$\mathcal{T}^e$}& \textbf{Fine-grained Textual Attributes}\\\hline
        
        AST & has\_name, has\_parameters, has\_body, has\_condition, has\_then\_body, has\_else\_body, has\_elif\_branch, has\_target, has\_value, contains \\ \hline
        
        CFG & sequential\_execution, true\_branch, false\_branch, alternate\_condition\_branch, condition\_evaluation, for\_loop\_body, for\_loop iteration\_range, while\_loop\_body, while\_loop\_condition, try\_block, exception\_handler, finally\_block, block\_exit, loop\_exit, loop\_back, break\_jump, condition\_false\_jump, function\_call \\ \hline
        
        DFG & contributes\_to, flows\_to \\ 
        \bottomrule
        
    \end{tabularx}
    \label{tab:textual_attributes}
\end{table}

\begin{table}[h]
    \centering
    \caption{Table of node types}
    \label{tab:node_types_classification}
    \vspace{-1em} 
    \begin{tabular}{@{}ll@{}} 
        \toprule
        $\mathcal{V}$ & $\mathcal{T}^v$ \\         \midrule
        Nodes & module, function\_definition, identifier, parameters, \\
              & default\_parameter, none, comment, block, try\_statement, \\
              & if\_statement, comparison\_operator, expression\_statement, \\
              & assignment, call, argument\_list, for\_statement, integer, \\
              & binary\_operator, subscript, pattern\_list, expression\_list, \\
              & while\_statement, parenthesized\_expression, string, \\
              & string\_start, string\_content, string\_end, elif\_clause, \\
              & else\_clause, augmented\_assignment, break\_statement, \\
              & continue\_statement, return\_statement, except\_clause, \\
              & finally\_clause \\
        \bottomrule
    \end{tabular}
\end{table}